%% file: top.tex
\documentclass[final]{cvpr}

\usepackage{times}
\usepackage{epsfig}
\usepackage{graphicx}
\usepackage{amsmath}
\usepackage{amssymb}

\usepackage{float}
\usepackage{booktabs}
\usepackage{caption}
\usepackage{subcaption}
\usepackage{multirow}
\usepackage{xcolor}

\mathchardef\mhyphen="2D
\usepackage[utf8]{inputenc}
\usepackage{comment}
\usepackage{tabularx,colortbl}
\usepackage{xparse,mathtools}
\usepackage{diagbox}
\usepackage{adjustbox,lipsum}

\usepackage{siunitx}
\usepackage{nopageno}

\usepackage[pagebackref=true,breaklinks=true,colorlinks,bookmarks=false]{hyperref}

\input{shortcuts}

\DeclareMathOperator{\E}{\mathbb{E}}

\def\cvprPaperID{778} %
\def\confYear{CVPR 2021}

\begin{document}

\title{Learning by Watching} %
\author{Jimuyang Zhang \quad Eshed Ohn-Bar\\
	Boston University\\
	{\tt\small \{zhangjim, eohnbar\}@bu.edu}
}

\maketitle

\input{sec_abstract}
\input{sec_intro}
\input{sec_related}

\input{sec_method}

\input{sec_results}

\input{sec_conclusion}

{\small
	\bibliographystyle{ieee_fullname}
	\bibliography{bibliography_custom}
}


\end{document}

%% file: shortcuts.tex
\newcommand{\ba}{\mathbf{a}}
\newcommand{\bB}{\mathbf{B}}

\newcommand{\bD}{\mathbf{D}}

\newcommand{\bl}{\mathbf{l}}
\newcommand{\bM}{\mathbf{M}}

\newcommand{\bR}{\mathbf{R}}
\newcommand{\bs}{\mathbf{s}}
\newcommand{\bT}{\mathbf{T}}

\newcommand{\bw}{\mathbf{w}}
\newcommand{\bx}{\mathbf{x}}

\newcommand{\btheta}{\boldsymbol{\theta}}

\newcommand{\btau}{\boldsymbol{\tau}}

\makeatletter
\DeclareRobustCommand\onedot{\futurelet\@let@token\@onedot}
\def\@onedot{\ifx\@let@token.\else.\null\fi\xspace}
\def\eg{e.g\onedot} 
\def\ie{i.e\onedot}

\def\etal{et~al\onedot}

\makeatother

\newcommand{\boldparagraph}[1]{\vspace{0.2cm}\noindent{\bf #1:}}

\definecolor{darkgreen}{rgb}{0,0.7,0}

%% file: sec_abstract.tex
\begin{abstract}
When in a new situation or geographical location, human drivers have an extraordinary ability to watch others and learn maneuvers that they themselves may have never performed. In contrast, existing techniques for learning to drive preclude such a possibility as they assume direct access to an instrumented ego-vehicle with fully known observations and expert driver actions. However, such measurements cannot be directly accessed for the non-ego vehicles when learning by watching others. Therefore, in an application where data is regarded as a highly valuable asset, current approaches completely discard the vast portion of the training data that can be potentially obtained through indirect observation of surrounding vehicles. Motivated by this key insight, we propose the Learning by Watching (LbW) framework which enables learning a driving policy without requiring full knowledge of neither the state nor expert actions. To increase its data, \ie, with new perspectives and maneuvers, LbW makes use of the demonstrations of other vehicles in a given scene by (1) transforming the ego-vehicle's observations to their points of view, and (2) inferring their expert actions. Our LbW agent learns more robust driving policies while enabling data-efficient learning, including quick adaptation of the policy to rare and novel scenarios. In particular, LbW drives robustly even with a fraction of available driving data required by existing methods, achieving an average success rate of 92\% on the original CARLA benchmark with only 30 minutes of total driving data and 82\% with only 10 minutes. 
\vspace{-0.6cm}
\end{abstract} 

%% file: sec_intro.tex
\section{Introduction}
Modern autonomous driving systems primarily rely on collecting vast amounts of data through a fleet of instrumented and operated vehicles in order to train imitation and machine learning algorithms~\cite{bansal2018chauffeurnet,bojarski2016end,Codevilla2019ICCV}. This process generally assumes direct knowledge of the sensory state of an ego-vehicle as well as the control actions of the expert operator. As driving algorithms largely depend on such costly training data to learn to drive safely, it is regarded as a highly valuable asset.

\begin{figure}[t]
    \centering
    \begin{tabular}{c}
    \hspace{-3mm}
      \includegraphics[trim=0.1cm 4.1cm 1.9cm 0.05cm,clip,width=3.26in]{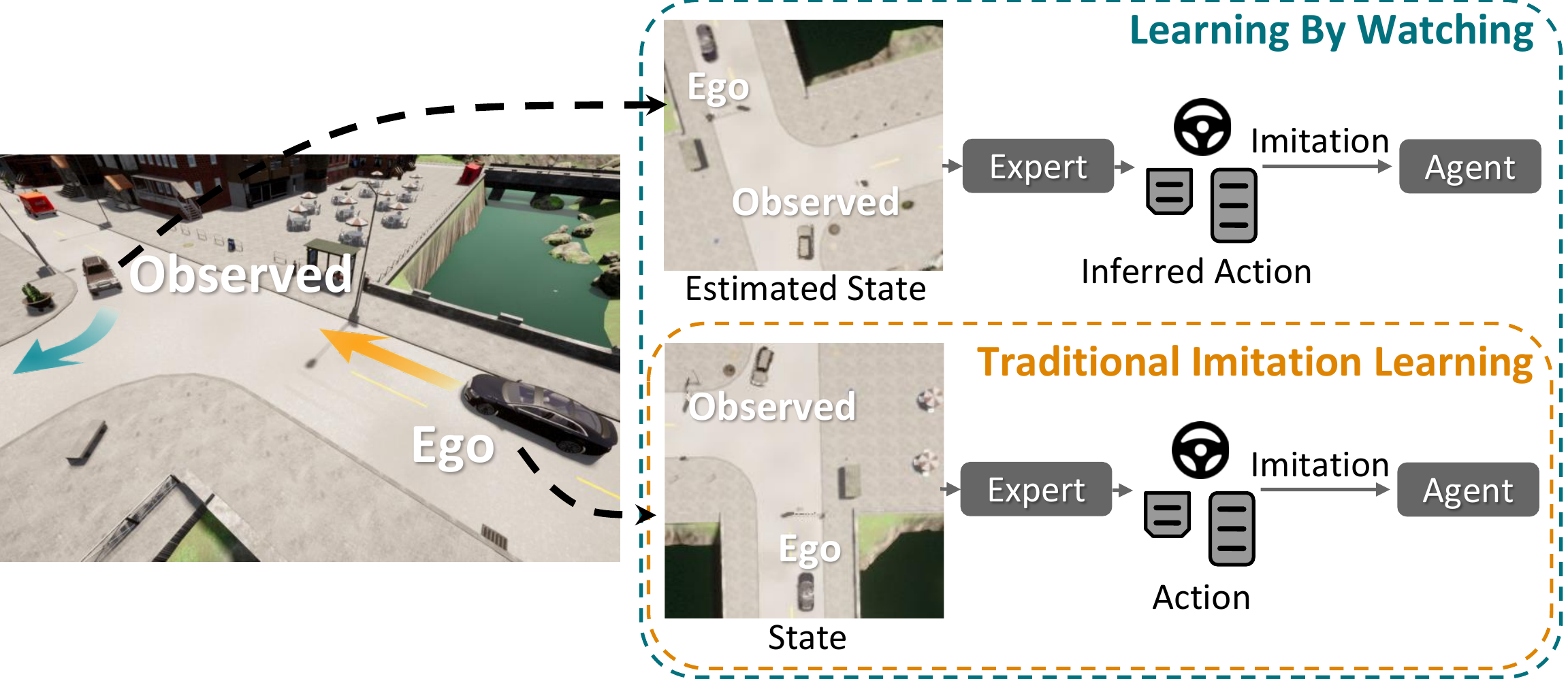} %
    \end{tabular}
    \vspace{-0.3cm}
    \caption{\textbf{Learning to Drive by Watching Others.} 
    While existing imitation learning approaches solely utilize data from an ego-vehicle perspective, our proposed Learning by Watching (LbW) framework learns a more robust and data-efficient policy from all available demonstration sources in a scene.
    } 
    \label{fig:f1}
    \vspace{-0.6cm}
\end{figure}

Unfortunately, such data requirements have resulted in self-captured data being amassed in the hands of a few isolated organizations, thereby hindering the progress, accessibility, and utility of autonomous driving technologies. First, while technological giants such as Uber, Tesla, Waymo and other developers may spend significant efforts instrumenting and operating their own vehicles, it is difficult to completely capture all of the driving modes, environments, and events that the real-world presents. Consequently, learned agents may not be able to safely handle diverse cases, \eg, a new scenario or maneuver that is outside of their in-house collected dataset. Second, while small portions of driving data may be shared, \ie, for research purposes, the bulk of it is predominantly kept private due to its underlying worth. Finally, we can also see how this current practice may lead to significant redundancy and thus stagnation in development. \textit{\textbf{How can we advance the underlying development process of safe and scalable autonomous vehicles?}}

In an application where the lack of a data point could mean the difference between a potential crash or a safe maneuver, we sought to develop a more efficient and shared paradigm to ensure safe driving. Towards this goal, we propose the LbW framework for learning to drive by watching other vehicles in the ego-vehicle's surroundings. Motivated by how human drivers are able to quickly learn from demonstrations provided by other drivers and vehicles, our approach enables leveraging data in many practical scenarios where \textit{\textbf{watched vehicles may not have any instrumentation or direct capture means at all.}} 

Towards more effective use of a collected dataset, LbW does not assume direct knowledge of either the state or expert actions. For instance, in the scenario shown in Fig.~\ref{fig:f1}, a non-instrumented human-driven vehicle or perhaps another company's autonomous vehicle is observed by our ego-vehicle as they turn and negotiate an intersection. LbW infers the observed agent's state and expert actions so that it may be used to teach our autonomous vehicle. By leveraging supervision from other drivers, our LbW agent can more efficiently learn to drive in varying perspectives and scenarios. The framework facilitates access to large amounts of driving data from human-driven vehicles that may not have been instrumented to directly measure and collect such data. 

While offering several benefits for scalability, learning a driving model via indirect means of watching surrounding vehicles in a scene also poses several challenges which we address in this work. Specifically, to advance the state-of-the-art of robust autonomous driving agents, we make the following \textit{\textbf{three contributions}}: (1) We propose LbW, a new paradigm which can help facilitate a more efficient development of driving agents, thus aiding real-world deployment, (2) we develop an effective two-step behavior cloning approach which infers the states and actions of surrounding vehicles without direct access to such observations, and (3) we validate the impact of LbW on the resulting driving policy through a set of novel experiments on the CARLA and NoCrash benchmarks. While previous approaches tend to exclusively focus on driving policy performance, \ie, requiring many hours of collected ego-vehicle training data, we instead emphasize the benefits of our approach by varying and limiting the amount of available data. We also demonstrate LbW to enable adaption to novel driving scenarios and maneuvers, without ever having direct access to an operator of an ego-vehicle performing such maneuvers.

%% file: sec_related.tex
\section{Related Work}
\label{label:relatedwork}

Our LbW framework builds on several recent advances in learning to drive, in particular imitation learning from intermediate representations, in order to utilize the demonstrations provided by all expert drivers in a given scene.

\boldparagraph{Imitation Learning for Autonomous Driving}
Over 30 years ago, Pomerleau~\cite{pomerleau1989alvinn} developed ALVINN, a neural network-based approach for learning to imitate a driver of an ego-vehicle. The approach requires ego-centric camera and laser observations and their corresponding operator steering actions. Since then, more elaborate approaches for learning to drive have emerged~\cite{liang2018cirl,zhu2017target,kober2013reinforcement,toromanoff2020end,chen2019attention}. Yet, techniques for imitation learning to drive are still widely employed due to implementation and data collection ease~\cite{bojarski2016end,osa2018algorithmic,li2018oil,zhao2019lates}, particularly the offline and supervised learning version of behavior cloning~\cite{Bain96aframework,pomerleau1989alvinn,muller2006off,ross2011reduction,Chen2015ICCVa}. Behavior cloning with conditional input is now a strong baseline for driving in urban settings~\cite{muller2018driving,liang2018cirl,Codevilla2018ICRA,Codevilla2019ICCV,ohn2020learning,lbc,rhinehart2018deep,rhinehart2018r2p2,filos2020can}. However, approaches generally require complete access to the perceptual observations of the vehicle together with the expert driver actions. Building on recent advances in imitation learning to drive, our work takes a step towards extending the traditional formulation so that it may be applicable to challenging cases where direct access to measurements cannot be assumed. Due to the difficulty of our learning by watching task, the approach hinges on the choice of underlying data representation, as discussed next.

\boldparagraph{Intermediate Representations for Autonomous Driving}
In parallel with steady progress in imitation learning to drive, the computer vision and machine learning communities have dedicated significant effort towards advancing perception and prediction tasks for autonomous vehicles. As pioneered by several researchers and developers, notably Dickmanns~\cite{dickmanns2007dynamic,dickmanns1992recursive,dickmanns1988dynamic,dickmanns2002development}, such tasks, \eg, semantic segmentation~\cite{cordts2016cityscapes}, 3D tracking of salient on-road objects from sensor input~\cite{Geiger2012CVPR,Zhoueaaw6661,yin2020center,Sauer2018CORL,8794224,muller2018driving,wang2019monocular,Fadadu2020multi,hawke2020urban,Chen2015ICCVa}, and future predictions~\cite{rudenko2020human,casas2020spagnn,choi2019looking,huang2019stgat,ivanovic2019trajectron,kosaraju2019social,liang2020pnpnet,sadeghian2019sophie,zhang2020stinet}, can be used in order to efficiently train a driving policy~\cite{Zhoueaaw6661,Behl2020label}. Of particular relevance to our study are approaches that leverage a Bird’s-Eye-View (BEV) of the scene~\cite{wang2019monocular,Fadadu2020multi,bansal2018chauffeurnet,Zeng2019end,schulter2018learning,gao2020vectornet}. While researchers may focus on accurately obtaining components of the BEV through a variety of sensor configurations~\cite{hong2019rules,houston2020one,roddick2020predicting,mani2020monolayout,reiher2020sim2real,lu2019monocular,hendy2020fishing,pan2020cross,philion2020lift,zhu2018generative,abbas2019geometric}, in our work we emphasize employing the BEV as a compact intermediate representation for learning a driving policy. Crucially, a BEV enables to efficiently transform observations between varying points of view and estimate agent-centric states when learning by watching. In contrast, transforming other types of representations across views, \eg,~\cite{coors2019nova,mildenhall2020nerf,zhu2018generative}, can be difficult for significantly differing perspectives.

\boldparagraph{Observational and Third-Person Imitation Learning} Several related recent studies tackle more general cases of imitation learning. For instance, methods may only require an observation of the state, such as video footage, without the underlying demonstrator actions.  Kumar~\etal~\cite{kumar2020learning} learns an inverse model that can generate pseudo-labels of actions for observed videos, and Murali~\etal~\cite{murali2015learning}~explores a similar learning by observation formulation for surgical tasks. Nait~\etal~\cite{Nair2017combining} presents a learning-based system using a pixel-level inverse dynamics model, \ie, for inferring actions that were taken by the human manipulator given a sequence of monocular images. However, such approaches still require access to the underlying state of the human expert, whereas we tackle the more general settings where direct access to neither the state nor the operator's actions is assumed. Another recent line of research analyzes third-person imitation learning~\cite{sermanet2018time,stadie2017third}, \ie, translating policies across slightly differing views of a shared workspace. Liu~\etal~\cite{liu2018imitation} present an imitation learning method based on a context translation model that can convert a demonstration from a third-person viewpoint (\ie, human demonstrator) to the first-person viewpoint (\ie, robot). However, these recent studies focus on toy settings and simplified visual environments, \eg, an inverted pendulum control task~\cite{stadie2017third}. In contrast, we are concerned with intricate multi-agent settings in the context of autonomous driving. Here, our task involves transforming and estimating demonstration data across drastically varying perspectives, frequent occlusion, dynamic scenes, and noisy agent tracking.

%% file: sec_method.tex
\begin{figure*}[!t]
    \centering
    \begin{tabular}{cccc}
        \includegraphics[trim=0cm 5.68cm 16.75cm 0cm,clip,width=1.55in]{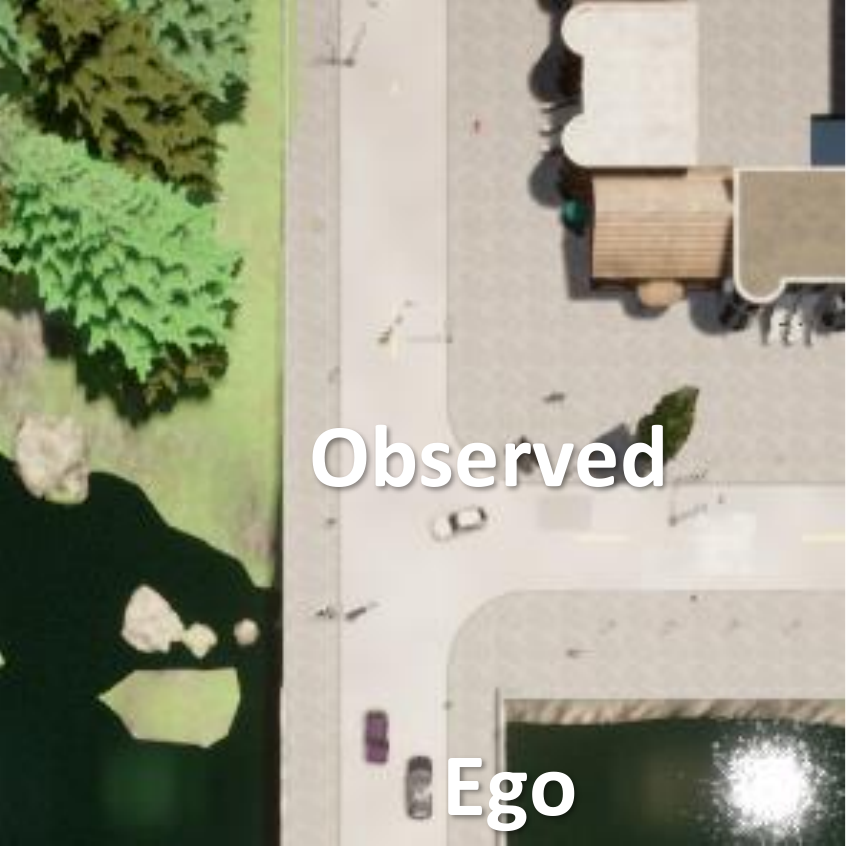} &
        \includegraphics[trim=0cm 0cm 0cm 0cm,clip,width=1.55in]{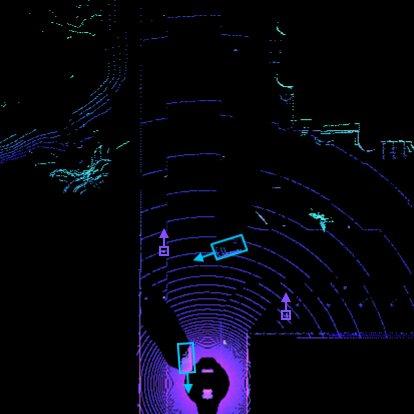} &
        \includegraphics[trim=0cm 0cm 0cm 0cm,clip,width=1.55in]{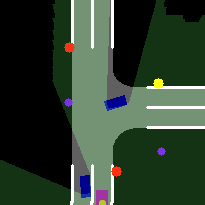} &
        \includegraphics[trim=0cm 0cm 0cm 0cm,clip,width=1.55in]{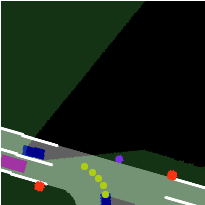}
        \\
        (a) Reference scene & (b) LiDAR map & (c) Ego-centric BEV & (d) Estimated BEV
    \end{tabular}
     \vspace{-0.2cm}
    \caption{\textbf{Bird's-Eye-View (BEV) Representation With Visibility Map Overlay.} We visualize (a) the reference top-down view of a scene, (b) its corresponding LiDAR map with vehicle and pedestrian detections, (c) the ego-centric BEV after integration with map information, including the ego-vehicle (magenta), surrounding vehicles (blue), pedestrians (purple circles), traffic lights (yellow, red), and visibility map (green overlay), and (d) the estimated BEV from the perspective of the observed vehicle. Dark yellow circles
    show future waypoints (Section~\ref{subsec:methodactions}) for the ego and observed vehicle. }
    \label{fig:visibility}
    \vspace{-0.1cm}
\end{figure*}

\section{Method}
\label{sec:method}

In this section, we describe our LbW approach. We first define the learning setting as it has some key differences compared to existing approaches for imitation learning to drive (Section~\ref{subsec:define}). Next, we describe our method for imitating the driving policies of not only our own instrumented and operated ego-vehicle, but of the surrounding vehicles as well. Our method consists of (1) an agent-centric input state representation that can be efficiently transformed to an observed agent frame of reference (Section~\ref{subsec:states}), (2) a visibility encoding for handling missing state information due to occlusion and significantly differing viewpoints (Section~\ref{subsec:states}), (3) a waypoint-based decomposition of the action representation aimed to efficiently learn from surrounding drivers (Section~\ref{subsec:methodactions}), and (4) a loss adjustment module for selectively refurbishing difficult and noisy samples inherent to the LbW task (Section~\ref{subsec:methodactions}). 

\subsection{Problem Definition} 
\label{subsec:define}
\boldparagraph{Imitation Learning Formulation} The objective of our driving agent is to generate a control action $\ba_t~\in~\mathcal{A}~=~[-1,1]\times[0,1]^2$, \ie, the steering, throttle, and brake at each time step $t$, in order to arrive to a pre-defined destination. The input to the agent is the current state $\bs_t \in \mathcal{S}$, which comprises of information about the vehicle and its surrounding context. The goal of the learning process is to learn a mapping
\begin{equation}
 \pi_{\btheta} \colon \mathcal{S} \to \mathcal{A}
\end{equation}
parameterized by $\btheta~\in~\mathbb{R}^d$, also known as the policy function. In this work, we are concerned with learning $\pi_{\btheta}$ from driver demonstrations. Traditionally, the process of imitation learning~\cite{osa2018algorithmic} assumes access to the state and expert of an operated ego-vehicle, \ie, trajectories of the form %
\begin{equation}
\label{eqn:sa}
    \btau^{expert} = \{\left(\bs_t, \ba^{*}_t\right)\}_{t=1}^{T}
\end{equation}
where $\ba^{*}_t$ are expert actions. The trajectories of directly experienced states and corresponding demonstrator actions are then used to learn a policy. The goal of the policy is to produce driving behavior that is similar to the behavior of the demonstrator. 

\boldparagraph{Imitation Learning by Watching} Despite ample related work studying our general decision-making problem, training LbW agents poses several unique challenges. Specifically, as in our case there is no direct access to either the state or action information of the watched agent, representing and inferring such information becomes a fundamental challenge\footnote{To clarify, it is only during LbW agent training that we have proxy and partial information regarding the states and expert actions of other agents. Once our LbW agent is trained, test-time driving is identical to the an agent learned via traditional ego-centric imitation learning, \ie, the state is directly observed.}. While surrounding vehicles are also driven by expert drivers, their trajectories are perceived indirectly from the viewpoint of the ego-vehicle. In this case, the underlying state and demonstrator actions must therefore be estimated, such that %
\begin{equation}
\label{eqn:lbwsa}
   \btau^{expert\mhyphen LbW} = \{\left(\hat{\bs}_t, \hat{\ba}^{*}_t\right)\}_{t=1}^{T}
\end{equation}
Due to the estimation task, training LbW agents is a very challenging problem. Specifically, such trajectory data must be properly represented, transformed, and inferred to avoid hindering driving performance. Next, we discuss how to effectively learn to drive from such potentially incomplete and noisy data.

\subsection{Estimating Agent-Centric States}
\label{subsec:states}
Estimating the agent-centric states of surrounding vehicles is key to using their demonstrations in Eqn.~\ref{eqn:lbwsa} to learn by watching. The complete state information is denoted as
\begin{equation}
\label{eqn:s}
    \bs_t = \left[ \bB_t, \bM_t, v_t, c_t \right]
\end{equation}
where $\bB_t$ is the current Bird's-Eye-View (BEV) image, $\bM_t$ is a visibility map of the scene computed from the viewpoint of the ego-vehicle, $v_t$ is the current speed, and $c_t$ is a categorical variable specifying a high-level navigation command following the goal-driven CARLA navigation task~\cite{Codevilla2018ICRA,lbc}. We further discuss these variables below.

\boldparagraph{Bird's-Eye-View State Representation} We propose to leverage a BEV image of the environment as it provides a compact and intuitive intermediate representation for learning a driving policy. The BEV representation will be crucial for addressing issues with differing viewpoints and missing observations when learning by watching. 

As shown in Fig.~\ref{fig:visibility}, our BEV is a rendered tensor $\bB~\in~\{0,1\}^{W\times H\times7}$ of a top-down view as perceived by the agent. The grid representation encodes each object type in a distinct channel, including the position of the road, lane marks, pedestrians, vehicles, and traffic lights and their state, \ie, red, yellow or green. The viewer position is fixed at the BEV's middle-bottom portion with a perpendicular orientation to the $x$-axis. Further details regarding BEV computation can be found in the supplementary.

While there has been significant work in obtaining BEV representations from a variety of sensor setups and maps, our work primarily focuses on leveraging such representations, in particular~\cite{yin2020center}, in order to learn a robust policy. We focus on the most essential state information for the BEV, such that other algorithms and more sophisticated representations, \eg, for agent dynamics and temporal context~\cite{hong2019rules,alahi2016social,casas2018intentnet} can be easily integrated to extend our analysis in the future. 

\boldparagraph{Observed Agent-Centric BEV} Observed drivers are reacting to the surroundings as perceived from their own perspective. The benefit of using a BEV representation, \ie, compared to low-level sensor information, is the crucial simplification it offers when estimating states from the perspective of other vehicles. 

To obtain an agent-centric BEV, we employ the 3D location $\bl$ and orientation $\alpha$ of an observed agent~\cite{yin2020center}. Next, any point $\bx \in \bB$ can be transformed through rotation and translation to an agent-centric frame of reference %
\begin{equation}
\label{eqn:transform}
   \bx_{\hat{\bB}} = \bR_{ego}^{observed}(\alpha) \bT_{ego}^{observed}({\bl}) \bx_{\bB}
\end{equation}
where $\bx$, written in homogeneous coordinates, is transformed to a new origin according to a rotation matrix calculated from $\alpha$ and a translation matrix calculated from $l$. This process produces a BEV for an observed vehicle $\hat{\bB}$.

In addition to the agent-centric BEV, we estimate the speed and high-level command of watched agents ($\hat{v}$, $\hat{c}$) by tracking and comparing their 3D position over time, as observed from the ego-vehicle perspective. We analyze motion of observed vehicles relative to their entrance position to an intersection in order to obtain $\hat{c}$.

\boldparagraph{Visibility Map} Our choice of a BEV state representation facilitates a more efficient learning by watching task. However, in practice, the viewpoint transformation in Eqn.~\ref{eqn:transform} can only roughly account for the state as it would have been perceived from the perspective of another vehicle. Due to occlusion and differing viewpoints in cases of complex urban settings, it is likely for the BEV estimation process to miss critical scene components. While missing scene context that could be perceived from a different perspective is a frequent issue in our experiments, Fig~\ref{fig:refurbishment} demonstrates a more severe example. Here, a stopped vehicle drastically limits the BEV of the ego-vehicle. This results in a poor estimate of the BEV for the observed vehicle, \ie, entirely missing an unobserved vehicle which greatly influences actions by the driver in the observed vehicle. In general, cases of occlusion can produce trajectories with various driving actions that cannot be well-explained by our estimate of the agent-centric BEV $\hat{\bB}$. 

We propose to alleviate this critical issue of potentially missing information by utilizing a visibility map representation~\cite{hornung2013octomap,hu2020you}. The visibility map, as shown in Figs.~\ref{fig:visibility} and~\ref{fig:refurbishment}, encodes areas in the BEV that were not directly perceived by the ego-vehicle. In this manner, the LbW agent can reason over crucial cues of unavailable information and learn to leverage the estimated BEV effectively during dense driving scenarios. While visibility and occupancy maps have been extensively studied in the general context of autonomous driving, its impact on learning policies from surrounding vehicles in complex, multi-agent settings with missing state information has not been previously analyzed. Moreover, it is important to note that we always compute the visibility map $\bM$ via ray-casting in the perspective of the ego-vehicle. This step ensures that samples with low visibility in the \textit{original frame of reference} are properly handled during LbW training. Eqn.~\ref{eqn:transform} is then used to obtain an agent-centric visibility map $\hat{\bM}$.

\subsection{Inferring Demonstrator Actions}
\label{subsec:methodactions}
When learning by watching, our goal is to leverage demonstrations provided by other drivers as additional data for training the policy function $\pi_{\btheta}$. However, the expert actions taken by drivers of other vehicles must first be inferred to be used as learning targets for behavior cloning approaches~\cite{Codevilla2018ICRA,Codevilla2019ICCV}. Nonetheless, inferring the steering, throttle, and brake control without having direct observation of such measurements is difficult. Therefore, as in the previous task of estimating agent-centric states, the LbW task hinges on the choice of a suitable underlying action representation. To effectively learn by watching, we propose to use a waypoint-based decomposition of the policy function.

\boldparagraph{Waypoint-Based Action Representation} The motion of watched vehicles as they are being driven alludes to the underlying operator actions. Motivated by this insight, our approach for inferring the actions of watched vehicles employs a representation based on waypoints. Each vehicle's observed motion can be unambiguously defined in terms of visited waypoints, \ie, a sequence of traversed positions by the expert in the BEV over time $\bw^{*} = \{\bw_0^{*},\bw_1^{*},...,\bw_K^{*}\}$~\cite{lbc}. During testing of the agent, the waypoints can then be passed as targets to a low-level controller $g$, \eg, a PID controller~\cite{visioli2006practical,lbc}. The controller can then produce the final policy, such that 
\begin{equation}
    \pi_{\btheta}(\bs) = g(f_{\btheta}(\bs))
\end{equation}
where $f_{\btheta}(\bs)$ is a learned waypoint prediction function. We do not use a parameter notation for $g$ as it does not include trainable parameters in our implementation. This decomposition can also simplify the overall policy learning task~\cite{kober2013reinforcement,muller2018driving}. 
While we can use a control model to infer low-level actions and use these as policy output targets directly, in practice we found this task to be ill-posed. Instead, the high-level waypoint-based action representation provides a standardized framework for combining action data from multiple different agents. It also enables the explicit handling of incomplete and noisy data inherent to the inferred experts, as discussed next.

\begin{figure}[t]
    \centering
    \begin{tabular}{cc}
        \includegraphics[trim=0cm 5.68cm 16.75cm 0cm,clip,width=1.5in]{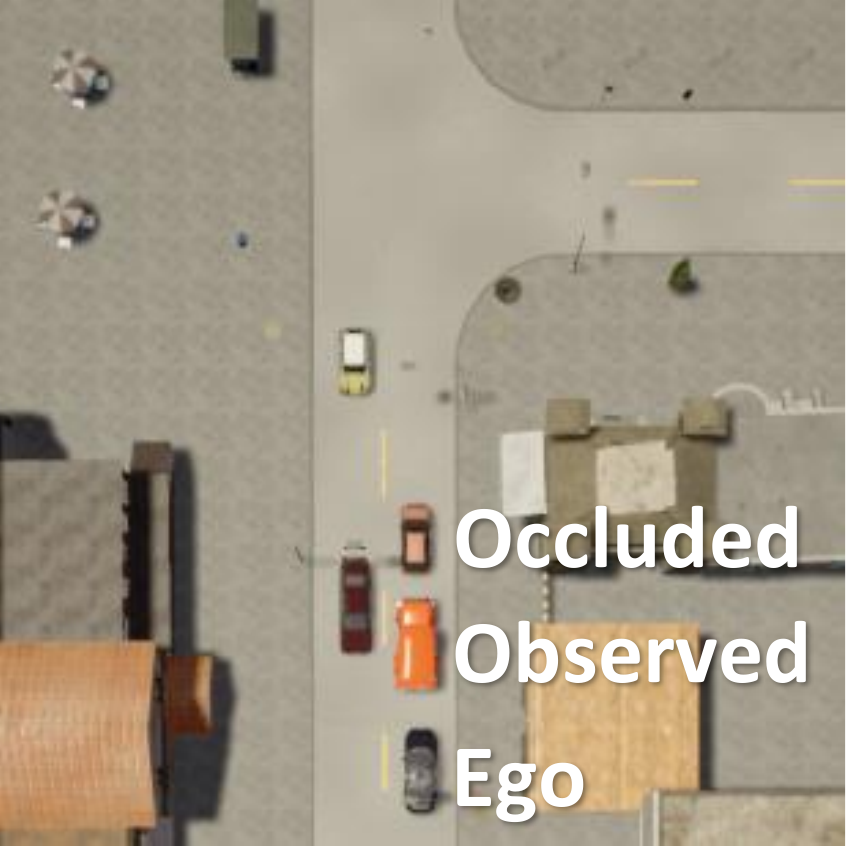}  & 
        \includegraphics[trim=0cm 0cm 0cm 0cm,clip,width=1.5in]{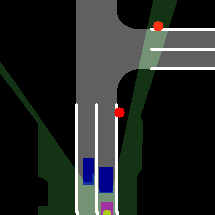}
        \\
        (a) Reference scene & (b) Ego-centric BEV  \\
        \includegraphics[trim=0cm 0cm 0cm 0cm,clip,width=1.5in]{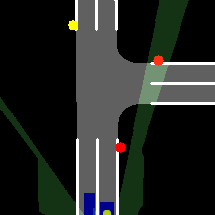} & 
        \includegraphics[trim=0cm 0cm 0cm 0cm,clip,width=1.5in]{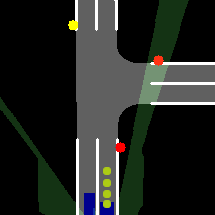}
        \\
        (c) Estimated BEV & (d) Refurbished waypoints
    \end{tabular}
    \vspace{-0.2cm}
    \caption{\textbf{Refurbishing Difficult Samples.} The example estimated view for an observed vehicle, shown in (c), is ambiguous due to occlusion. The proposed refurbishment process (Section~\ref{subsec:methodactions}) can correct the waypoint targets (d) to align with the estimated state and encourage proper agent behavior, \ie, to move forward and maintain a closer distance to the red light.     
    }
    \label{fig:refurbishment}
    \vspace{-0.25cm}
\end{figure}

\boldparagraph{Waypoint Refurbishment for Observed Vehicles} One of the main challenges of using the indirectly observed training data is the ability to learn from difficult samples with missing or noisy information. For instance, the observed waypoints may contain significant noise due to the underlying sensor setup and perception algorithm in Section~\ref{subsec:states}. Moreover, even in cases where the waypoints are observed accurately, the lack of state information to sufficiently explain their trajectory could potentially lead to ambiguity during agent training. In the visualized example of Fig.~\ref{fig:refurbishment}, the observed waypoints will not be fully explained by the available state information, \ie, expressing an inferred action to brake. Instead, the waypoints may vary arbitrarily despite an identical estimated state. Although the presence of numerous difficult and noisy samples could hinder policy training (Section~\ref{subsec:training}), discarding samples can remove potentially useful training data~\cite{song2019selfie}.

To take full advantage of the observed demonstrations, we employ a sample refurbishment process. We modify observed waypoints using a confidence hyper-parameter $\beta \in [0,1]$ for determining the amount of refurbishment, \ie,
\begin{equation}
\label{eqn:refub}
    \hat \bw^* = \beta \bw^* + (1-\beta) \hat{\bw}
\end{equation}
where $\bw^*$ are the observed waypoints and $\hat{\bw}$ are correction values. $\beta$ is used to specify the reliability of the waypoint
extraction step. As discussed further in Section~\ref{subsec:training}, the correction labels $\hat{\bw}$ are predicted from the estimated state $\hat{\bs}$ by an ego-only baseline model as it is trained using clean data, \ie, without any LbW data. In this manner, we can further mitigate noise and prevent overfitting to ambiguous or incorrect waypoint labels, thereby reducing their negative impact during training. We study the benefits of waypoint refurbishment for robust training with difficult cases of occlusion as well as overall perception noise in Section~\ref{sec:experiments}.

\subsection{Training}
\label{subsec:training}
\vspace{-0.25cm}
\boldparagraph{Dataset} Motivated by the success of behavior cloning algorithms,~\eg, \cite{Codevilla2019ICCV,Codevilla2018ICRA,lbc}, we pursue a supervised learning approach for learning from driver demonstrations. During training, as in previous work, we assume a collection of driver demonstrations comprising of sequences of ego-vehicle trajectories, $\bD^*_{ego} = \{ \btau_i^{expert} \}_{i=1}^{N}$. Differently from previous work, the proposed LbW method can then be used to extract \textit{new training data from the original ego-vehicle trajectories}. By leveraging human drivers which are operating the surrounding vehicles as expert demonstrators, we generate an additional trajectory for each agent $j$ that is observed throughout a drive $i$ such that $\bD^*_{LbW}~=~\{\hat{\btau}_{i,j}^{expert\mhyphen LbW} \}_{j=1}^{M}$, where $\hat{\btau}$ denotes refurbished trajectories. By incorporating the additional trajectories, our method enables learning robust policies with much smaller amounts of total collected driving time.

\boldparagraph{Robust Policy Learning} We learn a prediction function $f_{\btheta}$ by minimizing a waypoint-based behavior cloning (WBC) loss over samples from a dataset $\bD$
\begin{equation}
\label{eqn:loss}
\mathcal{L}_{\text{WBC}} = \E_{(\bs,\bw) \sim \bD} \left[ \ell_1 (\bw, f_{\btheta}(\bs)) \right] 
\end{equation}
The $\ell_1$-loss is computed over the next $K$ target waypoints that the autonomous vehicle should learn to predict. 

We train a robust LbW policy using Eqn.~\ref{eqn:loss} in three steps. First, we leverage $\bD^*_{ego}$ and train a baseline behavior cloning model, ${f}_{\hat{\btheta}}(\bs)$. In general, we assume that samples in $\bD^*_{ego}$ provide correct and clean supervision, as they were obtained through direct observation of driver actions. We then use the learned ego-only model to refurbish the LbW samples and obtain $\bD^*_{LbW}$, \ie, by setting $\hat{\bw} = {f_{\hat{\btheta}}}(\bs)$ in Eqn.~\ref{eqn:refub}. Thus, leveraging the baseline model to correct observed waypoints provides an explicit mechanism for removing overall noise and mislabeled samples. The refurbishment mechanism also aids in standardizing demonstrations across the various observed actors. Finally, the full model $f_{\btheta}$ is trained from scratch over the complete dataset $\bD^* = \bD^*_{ego}\cup\bD^*_{LbW}$. In this final step, the refurbished waypoint targets improve robustness by effectively adjusting the loss in Eqn.~\ref{eqn:loss}~\cite{song2019selfie}. During testing, we use our learned function $f_{\btheta}$ to map states to waypoints. The waypoints are then given to the PID controller to obtain the policy $\pi_{\btheta}$, \ie, as an inverse dynamics model for mapping waypoints to actions.

\subsection{Implementation Details}
\label{subsec:implement}
\vspace{-0.15cm}
We implement $f_{\btheta}$ as a convolutional neural network with trainable parameters $\btheta$. We follow Chen~\etal~\cite{lbc} and train a state-of-the-art BEV-based conditional behavior cloning agent (see supplementary for details). 

\boldparagraph{Visibility Map Integration} We are interested in analyzing the impact of integrating the visibility with the BEV on policy learning as means to encode missing state information and reason over potentially ambiguous corresponding waypoint labels. As visibility maps have not been thoroughly studied in this specific context, we explore two network structures for integrating visibility cues, early and late fusion. In the early fusion scheme, the visibility map is leveraged as another channel concatenated to the BEV prior as an input to the backbone. However, the type of information encoded in the visibility map significantly differs in its purpose from the type of object position information that is found in the BEV. Therefore, we also experiment with a late fusion scheme, with a separate backbone for the visibility map. Both fusion schemes enable the network to weigh the two sources of information as needed, \ie, when waypoints are poorly explained by the BEV alone. 
\vspace{-0.17cm}

%% file: sec_results.tex
\begin{table*}[!t]
    \centering
        \caption{\textbf{Ablation Study.} Comparison of driving success rate (\%) for the proposed approach (LbW) with various visibility fusion schemes. The baseline model (Ego) is trained by traditional behavior cloning. Mean and standard deviation are shown over three runs using the original CARLA benchmark (OB) and the NoCrash benchmark (Regular: NC-R, Dense: NC-D).}
        \label{tab:unprivileged}
        \vspace{-0.2cm}
    \noindent\adjustbox{max width=\textwidth}{
    \begin{tabular}{l c| c || c c c |c|| c c c | c || c}
    \toprule
    & \multicolumn{3}{c}{\textbf{One Hour}} & & \multicolumn{3}{c}{\textbf{30 Minutes}} & & \multicolumn{3}{c}{\textbf{10 Minutes}} \\
    \cmidrule{2-4}  \cmidrule{6-8} \cmidrule{10-12}
      &  NC-R & NC-D & OB & & NC-R & NC-D & OB & & NC-R & NC-D & OB \\
    \midrule
    Ego (Baseline)  &  $46 \pm 1$  & $18 \pm 1$ &  $56 \pm 1$ &&  $26 \pm 2$  & $12 \pm 1$ & $68 \pm 1$  && $24 \pm 1$& $0 \pm 0$ & $64 \pm 1$ \\
    LbW  &  $64 \pm 1$  & $24 \pm 1$ &  $74 \pm 1$ && $52 \pm 0$ &  $\textbf{24} \pm \textbf{0}$ & $68 \pm 1$  && $34 \pm 1$ & $6 \pm 1$ & $\textbf{82} \pm \textbf{4}$\\
    LbW + Visibility (Early)  &  $52 \pm 1$  & $\textbf{24} \pm \textbf{0}$ &  $76 \pm 1$ && $54 \pm 1$ &  $18 \pm 1$ & $72 \pm 3$ && $28 \pm 1$ & $6 \pm 1$ & $64 \pm 1$\\
    LbW + Visibility (Late) &   $\textbf{92} \pm \textbf{3}$ & $\textbf{24} \pm \textbf{0}$ &  $\textbf{92} \pm \textbf{1}$  && $\textbf{74} \pm \textbf{2}$&  $\textbf{24} \pm \textbf{0}$ & $\textbf{92} \pm \textbf{0}$  && $\textbf{52} \pm \textbf{1}$ & $\textbf{20} \pm \textbf{2}$ & $68 \pm 1$ \\
    \bottomrule
    \end{tabular}}
  \vspace{-0.25cm}
\end{table*}

\section{Experiments}
\label{sec:experiments}
We use the 0.9.9 version of the CARLA simulator~\cite{Dosovitskiy2017CORL} for generating diverse multi-agent driving scenes and evaluating the proposed LbW framework.  

\boldparagraph{Original and NoCrash CARLA Benchmarks} The CARLA benchmarks~\cite{Dosovitskiy2017CORL,Codevilla2018ICRA,Codevilla2019ICCV} employ Town 1 of the simulation for training and Town 2 for testing under varying navigation tasks. The original benchmark is now obsolete~\cite{ohn2020learning,lbc}, yet previous analysis used older versions of the simulation (0.8.4-0.9.6). To analyze performance under more recently introduced functionalities, \eg, complex pedestrian behavior and changed graphics, we keep this benchmark for reference and only use the most challenging task of dynamic navigation settings~\cite{Dosovitskiy2017CORL}. We also consider the more challenging NoCrash benchmark settings~\cite{Codevilla2019ICCV}, which employ various traffic density conditions (\eg, regular, dense) while not allowing for collisions. Previous approaches on CARLA generally require large amounts of training data, specified as ego-vehicle driving hours, \eg, between five hours in~\cite{lbc} to 10 hours in~\cite{Codevilla2019ICCV}. In contrast, our proposed method can effectively utilize much smaller datasets. Therefore, we focus on much lower total driving time of between 10 minutes and up to one hour. This experimental setup also ensures that our method is beneficial to low-data settings, \eg, for rare maneuvers.

\begin{figure}[t]
    \centering
    \begin{tabular}{cc}
        \includegraphics[trim=0cm 0cm 0cm 0cm,clip,width=1.45in]{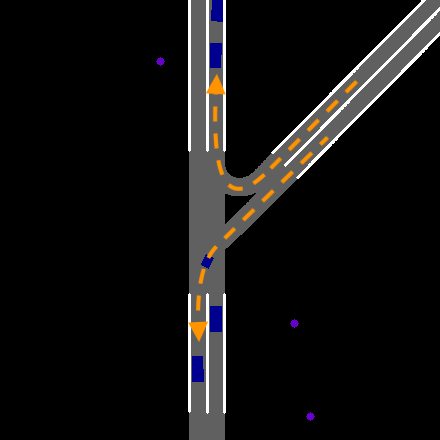} &  
        \includegraphics[trim=0cm 0cm 0cm 0cm,clip,width=1.45in]{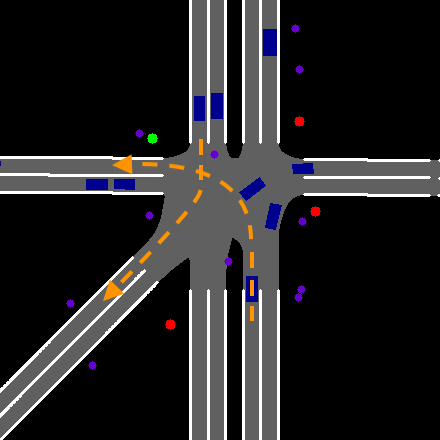}
    \end{tabular}
   \vspace{-0.25cm}
    \caption{\textbf{Adaptation Benchmark.} We visualize example intersections from Town 3 together with agent testing routes (not seen in training) in orange arrows. 
    }
    \label{fig:f3}
   \vspace{-0.3cm}
\end{figure}

\boldparagraph{Adaptation Benchmark} To further analyze the data-efficiency limitations of our approach, we also introduce a new benchmark defined by training over Town 1 and adapting the model to the most difficult town in the simulator, Town 3. The commonly used first two towns are relatively small and are comprised of similar features, \eg, simple two-lane roads with only orthogonal three-way intersections. In contrast, Town 3 has diverse multi-lane roads and intersections, with roundabouts, five-way intersections, and diagonal and curved turns. For adaptation, we define a disjoint set of maneuvers for training and testing and collect a small amount of data (a total of 10 minutes driving time) to be used for adapting the model trained in Town 1. Example test routes are shown in Fig.~\ref{fig:f3}. We define a CARLA and NoCrash-type benchmarks on Town 3 with various driving conditions. These challenging settings aim to analyze a realistic scenario where an autonomous vehicle must quickly learn to operate in a new situation or geographical location. 

\subsection{Results}
We conduct four main experiments to analyze the benefits of different components in our proposed approach. First, we begin by comparing the performance of the LbW agent with the traditional behavior cloning baseline~\cite{lbc}. Second, we analyze the various underlying network structures for fusion of the visibility map. Third, we explore the performance of the LbW agent under a low-data regime. Finally, we investigate the impact of the proposed label refurbishment process when training robust driving agents.

\boldparagraph{LbW vs. Baseline} We first discuss our results using one hour of total driving data. The analysis in Table~\ref{tab:unprivileged} shows the driving performance, in terms of success rate, for an agent trained with our proposed approach. Results are shown for both the NoCrash (NC) regular and dense benchmarks as well as the reference original CARLA benchmark (OB). With one hour of training data, we find the proposed LbW approach to consistently improve overall driving performance, \ie, by $39\%$, $33\%$, and $32\%$ over the baseline agent for the NC-Regular, NC-Dense, and original CARLA settings, respectively. Notably, the consistent improvements are achieved by the LbW agent prior to integrating the visibility map or refurbishment step. We emphasize that the driving dataset collected for training both the LbW and baseline models is identical. Yet, the baseline model only leverages demonstration data that has been directly observed by the ego-vehicle. In contrast, through effective cross-perspective reasoning, our method increases the available training samples with respect to the number of surrounding agents observed. As such, the baseline’s performance can match LbW asymptotically in principle, \ie, through exhaustively driving in various scenes, maneuvers, and perspectives. This is inefficient and costly in practice.

\boldparagraph{Integrating the Visibility Map} We sought to understand the benefits of different model choices in our learning by watching framework. Here, we analyze the underlying network structure for integrating the visibility map with respect to the ultimate driving task. Our findings with one hour of training data highlight the importance of effectively integrating visibility cues with BEV-based cues. Specifically, we achieve a $92\%$ success rate for the late fusion-based LbW agent compared to $64\%$ obtained by an agent without a visibility map on NC-Regular. We also nearly solve the original CARLA benchmark, achieving a $92\%$ success rate. Early fusion of the visibility map is less successful, as our visibility cues differ in their function compared to the positional BEV cues. Thus, our model benefits from an architecture with distinct processing for the two types of cues. It is useful to contrast this finding with the 3D object detection study of Hu~\etal~\cite{hu2020you}, where an early fusion scheme was found to work best. One reason for this discrepancy could be due to the differences in the underlying learning task. In this study, we utilize the visibility cue to encode a more complex notion of missing cross-perspective information when training a driving policy.

\boldparagraph{Data-Efficiency in Training} Our primary goal is to enable researchers to easily train robust driving agents while minimizing operation and data collection costs. Towards this goal, in this experiment we deliberately consider an often neglected aspect of data-efficiency when training driving policies. Here, to highlight the benefits of our LbW approach, we use a much smaller amount of driving data than is generally considered, \eg, 30 and 10 minutes in total. This is a challenging experiment given that our models are initialized from scratch. Our key finding in Table~\ref{tab:unprivileged} reveals how \textit{\textbf{our LbW agent outperforms the baseline model trained with six times the amount of data}}. For instance, for NC-Regular test settings, our agent trained with 10 minutes of data achieves a $52\%$ success rate compares to $46\%$ with traditional behavior cloning and one hour of data.

\begin{table}[!t]
    \centering
        \caption{\textbf{Adaptation Results.} Quantitative results for policy adaptation to novel maneuvers on Town 3.}
        \label{tab:town3}
        \vspace{-0.15cm}
    \noindent\adjustbox{max width=\columnwidth}{
    \begin{tabular}{l c | c || c}
    \toprule
    &  NC-R & NC-D & OB\\
    \midrule
    Ego (Baseline) & $40 \pm 0$ & $40 \pm 0$ & $60 \pm 0$ \\
    
    LbW & $60 \pm 1$ & $\textbf{60} \pm \textbf{0}$ & $80 \pm 1$ \\
    
    LbW + Visibility (Early) & $60 \pm 1$ & $60 \pm 1$ & $60 \pm 1$\\
    
    LbW + Visibility (Late) & $\textbf{100} \pm \textbf{0}$ & $40 \pm 0$ & $\textbf{100} \pm \textbf{0}$\\
    \bottomrule
    \end{tabular}}
   \vspace{-0.25cm}
\end{table}

\begin{figure}[!t]
    \centering
    \begin{tabular}{cc}
        \includegraphics[trim=0cm 0cm 0cm 0cm,clip,width=1.45in]{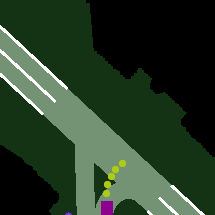} &  
        \includegraphics[trim=0cm 0cm 0cm 0cm,clip,width=1.45in]{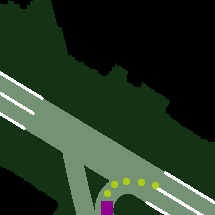}
        \\
        (a) Ego (Baseline) &  (b) LbW (Proposed)
    \end{tabular}
    \vspace{-0.2cm}
    \caption{\textbf{Qualitative Adaptation Results.} While both agents leverage the same short adaptation route, the LbW agent effectively learns entirely new maneuvers solely through watching them performed by other vehicles. LbW results in more accurate and safe predicted waypoints.
    }
    \label{fig:town3_result}
    \vspace{-0.4cm}
\end{figure}

\boldparagraph{Data-Efficiency in Adaptation} We further explore the practical limitations of our proposed approach by adapting the driving agent to a town with completely different layout characteristics and maneuvers. In this realistic scenario, the agent encounters an environment differing from its original dataset and must efficiently adapt to the new environment in order to avoid an unsafe maneuver, \eg, as visualized in Fig.~\ref{fig:town3_result}. Here, even a single trajectory watched at an intersection can potentially teach the ego-vehicle an entirely new maneuver, \ie, turning at a five-way intersection. As not utilizing such critical data would lead to poor driving performance on the most complex CARLA town, these settings provide an ideal test case. Table~\ref{tab:town3} shows how the baseline model generally struggles to adapt to the new environment. In contrast, our LbW agent can achieve up to $100\%$ success rate on NC-Regular driving settings, solely through watching novel maneuvers being performed by other vehicles. As shown in Fig.~\ref{fig:town3_result}, the LbW adaptation process results in improved waypoint predictions for the novel maneuvers. Nonetheless, the conditions on this new town, especially during dense settings, can create drastically different visibility patterns compared to the ones in Town 1. Hence, we find that the adaptation of our LbW agent without visibility cues is more efficient for this difficult task, motivating future work in learning generalized driving policies. 

\boldparagraph{Robust Policy Learning} Given the analysis on the underlying network structure and choice of late fusion scheme, we now quantify the role of waypoint refurbishment when handling mislabeled samples inherent to LbW datasets. In particular, we sought to further analyze our proposed approach on more general settings, for instance, settings with a lower quality sensor, algorithm, or map, as such noise can impact the learning by watching process. To ensure our findings are relevant across various sensor and algorithm configurations, we further add waypoint noise into our one hour dataset when training our models. By improving robustness, \ie, through adjustment of the training loss, we show refurbishment to consistently improve driving success rates in Table~\ref{tab:refurbishment}. Refurbishment leads to further gains of $25\%$, $13\%$, and $11\%$ on the NC-Regular, NC-Dense, and original benchmark settings, respectively.

 \begin{table}[!t]
     \centering
     \caption{\textbf{Refurbishment Analysis.} Impact of waypoint refurbishment on driving success rate (\%). Results are shown using the late fusion visibility integration scheme. 
      \vspace{-0.05cm}
         }
         \label{tab:refurbishment}
         \vspace{-0.2cm}
     \noindent\adjustbox{max width=\columnwidth}{
     \begin{tabular}{l c | c || c}
     \toprule
      Town 2 & NC-R & NC-D & OB \\
      \midrule
     LbW + Visibility   & $ 64\pm 0$  & $ 32\pm3 $ &  $ 86\pm 1$ \\
     LbW + Visibility (Refurbishment)  & $ \textbf{80} \pm \textbf{1}$  & $ \textbf{36}\pm \textbf{0}$ &  $ \textbf{96}\pm\textbf{0} $ \\
     \midrule
     Town 3 & NC-R & NC-D & OB \\
     \midrule
     LbW + Visibility   & $ 40\pm 0$  & $ 30\pm 1$ &  $ \textbf{60}\pm \textbf{0}$ \\
     LbW + Visibility (Refurbishment)  & $ \textbf{60} \pm \textbf{0}$  & $ \textbf{40}\pm \textbf{0}$ &  $ \textbf{60}\pm\textbf{0} $ \\
     \bottomrule
     \end{tabular}}
     \vspace{-0.35cm}
 \end{table}

%% file: sec_conclusion.tex
\section{Conclusion}
In this work, we have presented the learning by watching framework. Based on our experiments, we showed that leveraging supervision from inferred state and actions of surrounding expert drivers can lead to dramatic gains in terms of driving policy performance, generalization, and data-efficiency. While our approach enables learning from incomplete data and noisy demonstrations, an important next step would be further validation under such realistic challenges. For instance, although existing approaches on CARLA generally assume expert optimality, this assumption may not hold in practice when learning to drive by watching diverse and imperfect drivers. The complementarity of the LbW approach to additional sample-efficient imitation learning approaches, \eg,~\cite{blonde2019sample,jeon2018bayesian,ho2016generative}, could also be studied in the future. More broadly, we hope that our work motivates others to pursue a more shared and efficient development of autonomous vehicles at scale.